\documentclass{article}
\usepackage{PRIMEarxiv}
\usepackage[utf8]{inputenc}
\usepackage[T1]{fontenc}
\usepackage{nicefrac}
\usepackage{microtype}
\usepackage{fancyhdr}
\usepackage{natbib}
\setcitestyle{authoryear,round,citesep={;},aysep={,},yysep={;}}
\newcommand{\method}{\textsc{LEMMO}}
\usepackage{amsmath,amsfonts}
\usepackage{graphicx}
\usepackage{hyperref}
\usepackage{url}
\usepackage{booktabs}
\usepackage{pifont}
\usepackage{xcolor}
\usepackage{multirow}
\newcommand{\cmark}{\textcolor{green!60!black}{\ding{51}}}
\newcommand{\xmark}{\textcolor{red}{\ding{55}}}
\title{LEMMO: A Large Electromagnetic Model leveraging Pretrained LLMs for Million-Sample Signal Understanding}

\author{
\textbf{Mingkai Li}\thanks{Equal contribution} \quad
\textbf{Jiajun Wu}\footnotemark[1] \quad
\textbf{Yihan Mao \quad Zilin Sang \quad Taosong Yu \quad Yuan Ma \quad Minghe Liu}\\
\textbf{Qianqian Yang\thanks{Corresponding Author} \quad Zhiguo Shi}\\
College of Information Science and Electronic Engineering \\
Zhejiang University \\
\texttt{\{limingkai,wujiajun,qianqianyang20,shizg\}@zju.edu.cn}\\
}

\date{}
\begin{document}

\maketitle

\begin{abstract}
Large language models have expanded beyond text to vision,
audio, and other modalities, but understanding raw
electromagnetic (EM) signals remains challenging.
Wideband recordings can contain millions of samples, whereas
existing waveform-based EM models are primarily evaluated
on short in-phase and quadrature (IQ) segments, leaving long-recording understanding
and detailed description insufficiently explored.
We introduce \method{}, a large EM model for
instruction-driven recognition and description of long
raw IQ recordings.
\method{} combines a linear-attention EM encoder with
anti-aliased temporal resampling to process up to eight
million complex samples. Its EM--text interaction module
conditions learnable queries on instructions before
retrieving signal evidence. Compact queries support
recognition, while structured queries with auxiliary
factual supervision organize scene-level properties,
signal attributes, and cross-signal relationships for
detailed description. We also construct and release a
unified 1.38-TB corpus of real-world and synthetic EM
signals, including LEMMO-ShortQA and LEMMO-LongDesc.
Compared with MERLIN-LoRA, \method{} improves mean
EM-Bench perception accuracy by 15.77 \%,
mean strategy-generation ROUGE-L F1 by 4.38 points,
and mean LEMMO-ShortQA accuracy by 14.31 \%.
It also improves categorical and numerical fact F1 on
both LEMMO-LongDesc tasks while preserving its language
backbone's MMLU performance. Our code and datasets are
available at \url{https://github.com/N1rv4n4/LEMMO} and
\url{https://huggingface.co/datasets/DaftJun/RFLAS}, respectively.
\end{abstract}

\section{Introduction}
\label{sec:introduction}


Electromagnetic (EM) signals carry information about the wireless
activities around us: which devices are transmitting, how they use
the spectrum, and how their transmissions overlap or interfere.
Interpreting this information supports applications across wireless
communications, environmental sensing, autonomous systems, and
security~\citep{zhang2024spectrum,dong2021sr2cnn,yu2024openset}.
Such understanding extends beyond isolated waveform classification
to identifying signal components, characterizing their behavior,
and interpreting their temporal and spectral relationships.
Large language models (LLMs) have expanded beyond language to
modalities such as vision, enabling natural-language interaction
with increasingly diverse types of information \cite{dai2023instructblip,huang2023language,chen2024internvl}. Extending this
capability to EM signals requires a corresponding interface that
extracts instruction-relevant evidence from raw waveforms and
incorporates it into response generation.



Existing research provides important steps toward this objective.
Task-specific deep learning methods have advanced automatic
modulation recognition (AMC)~\citep{o2018over,li2023expert}, jamming pattern
recognition (JPR)~\citep{wang2019jamming,wang2026jamming}, and radio frequency (RF)-based unmanned aerial vehicle (UAV) identification~\citep{yu2025smnet,cai2024uav}. EM foundation models, including
SpectrumFM and EMind, broaden this scope by learning transferable
representations through self-supervised pretraining~\citep{zhou2025spectrumfm,luo2025emind}. MERLIN and PReD
further connect signal representations with LLMs for question answering and strategy generation~\citep{Shen_2026_CVPR,han2026pred}.
However, representative waveform-based models are primarily trained
and evaluated on segments containing hundreds to a few thousand in-phase and quadrature (IQ)
samples; these evaluations do not establish their ability to
understand million-sample raw-IQ recordings~\citep{zhou2025spectrumfm,luo2025emind,Shen_2026_CVPR}.
RF-GPT explores wideband understanding through spectrograms, but
magnitude spectrograms do not explicit retain phase information,
and their time-frequency detail depends on image
resolution~\citep{zou2026rf}.

\begin{table*}[t]
\centering
\caption{Comparison of capabilities between existing EM models and LEMMO.}
\label{tab:model_comparison}

\begin{tabular}{lccccc}
\toprule
\multirow{2}{*}{Model} &
Raw IQ &
Long IQ &
Natural &
Detailed Signal &
UAV RF \\
&
Encoding &
Sequence &
Language QA &
Description &
Recognition \\
\midrule
SpectrumFM  & \cmark & \xmark & \xmark & \xmark & \xmark \\
EMind             & \cmark & \xmark & \xmark & \xmark & \xmark \\
RF-GPT           & \xmark & \xmark & \cmark & \cmark & \xmark \\
PReD              & \xmark & \xmark & \cmark & \xmark & \xmark \\
MERLIN          & \cmark & \xmark & \cmark & \xmark & \xmark \\
\midrule
\textbf{LEMMO}                & \textbf{\cmark} & \textbf{\cmark} & \textbf{\cmark} & \textbf{\cmark} & \textbf{\cmark} \\
\bottomrule
\end{tabular}

\end{table*}

We study instruction-driven understanding directly from continuous
wideband IQ recordings spanning multiple
frequency channels at a fixed receiver center frequency.
The temporal locations and frequency bands of signals of interest
are unknown in advance, so the model cannot rely on prior
target-specific cropping, bandpass filtering, or downsampling.
This setting presents two coupled challenges.
First, high sampling rates produce long inputs even over brief
intervals: a 10\,ms recording at 100\,MS/s contains one million
complex samples. Processing these inputs requires efficient
encoding that preserves local waveform characteristics and
integrates context across the recording.
Second, the evidence needed from a recording depends on the
instruction. Recognition requires category-discriminative features,
whereas detailed description requires distinguishing signal
components, associating attributes with them, and describing their
relationships. The signal-to-language interface must therefore
both select relevant evidence and organize it for the requested task.
To address these challenges, we introduce \method{}, a framework
that combines hierarchical waveform processing with
instruction-conditioned evidence retrieval.
A causal linear-attention EM encoder extracts signal features,
and a Temporal Resampler reduces the sequence length for recordings
exceeding one millio complex samples.
The resampler compresses encoded subsequences by a factor of 64,
concatenates them in temporal order, and integrates their context
through bidirectional linear attention; shorter recordings bypass
resampling. This design supports inputs of up to eight million
complex samples. The EM-text interaction module then conditions
learnable queries on the instruction before they attend to
signal features, making evidence selection instruction-dependent.
The retrieved representations are projected into a pretrained
LLM's embedding space for response generation.

We further tailor the query organization and supervision to the
different requirements of recognition and description.
\textbf{Q16 Recognition} uses 16 general-purpose queries trained
jointly for language generation and classification.
\textbf{Q128 Description} uses 128 queries organized into global,
object-feature, and relation groups, together with physical position
information and auxiliary factual supervision.
This structure organizes evidence at the scene, object, and
relation levels, while the auxiliary objectives supervise the
representations used for generation without passing fact-head
predictions to the LLM.To support training and evaluation, we construct and release a
unified corpus of real-world and synthetic IQ recordings, including
LEMMO-ShortQA for recognition and LEMMO-LongDesc for detailed
description. Across the recognition tasks evaluated in EM-Bench
and LEMMO-ShortQA, \method{} achieves higher mean accuracy than
Merlin-LoRA.

Our contributions are threefold:



\begin{itemize}
    \item We combine linear-attention waveform encoding, temporal
    compression, and cross-subsequence contextualization to
    support inputs of up to eight million complex samples.
    \item We condition learnable queries on instructions before signal retrieval and develop two complementary configurations:
    compact recognition queries with discriminative supervision,
    and structured description queries with auxiliary factual
    supervision.
    \item We release real-world and synthetic raw-IQ recordings for
    representation learning and language supervision, including
    LEMMO-ShortQA and LEMMO-LongDesc for recognition and detailed
    descriptions of signal components, attributes, and relationships.
\end{itemize}

\section{Related Work}
\label{sec:related_work}

Research on EM signal understanding follows two directions: EM foundation models learn transferable signal representations, while EM-language models connect signal with natural language.

\subsection{EM Foundation Models}

EM foundation models use self-supervised learning to extract
transferable representations from unlabeled signal
data~\citep{shi2023gaf,kong2023transformer,xiao2024mclhn,davaslioglu2022self}.
Their inputs and learning objectives vary. RIS-MAE applies
masked reconstruction to raw IQ signals, while IQFM uses
contrastive learning on IQ
streams~\citep{liu2025ris,mashaal2026iqfm}.
Other efforts, including SpectrumFM, 6G WavesFM, and WiFo,
explore representation learning for spectrum management,
communications, sensing, localization, and channel
prediction~\citep{zhou2025spectrumfm,aboulfotouh20256g,liu2025wifo}. Recent studies also improve the efficiency and flexibility
of signal encoding. Patch-based tokenization reduces sequence
length, facilitating the processing of inputs containing
thousands of samples~\citep{cheraghinia2025unified}.
EMind introduces length-adaptive multi-signal packing to
accommodate heterogeneous EM
inputs~\citep{luo2025emind}.
These approaches broaden the range of signals that a common
encoder can process.

However, representation learning alone does not provide
instruction-driven interaction or natural-language description.
Moreover, representative models are evaluated primarily on
inputs containing a few thousand
samples~\citep{luo2025emind,cheraghinia2025unified},
which cover only brief observation intervals at high sampling
rates. Such intervals may exclude recurring bursts, frequency
changes, or relationships between temporally separated events.
LEMMO addresses this limitation through hierarchical processing
of raw IQ recordings containing up to eight million samples,
coupled with an instruction-conditioned language interface.

\subsection{EM-Language Models}

EM-language models integrate signal information with LLMs
in different ways. RadioLLM uses prompt and token
reprogramming for radio-signal
processing~\citep{chen2025radiollm}, while WirelessLLM
fine-tunes LLMs for wireless communication
tasks~\citep{10582827}. MERLIN jointly learns signal
representations and adapts a pretrained LLM for EM
perception and strategy generation~\citep{Shen_2026_CVPR}.

Other approaches transform signals into visual or symbolic
representations. Seeing Radio and RF-GPT use image-based
signal representations for recognition and text
generation~\citep{zou2026seeing,zou2026rf}.
PReD integrates multiple signal views, including waveforms,
spectra, time-frequency representations, and constellation
diagrams, to support EM perception, recognition, and
decision-making~\citep{han2026pred}.
DiSC-AMC discretizes numerical signal features into symbolic
tokens and combines them with nearest-neighbor retrieval
for modulation classification~\citep{rostami2026largelanguagemodelsperform}.

These approaches enable language-based EM processing but
differ in the signal information they preserve.
Visual representations capture selected signal characteristics,
with fidelity determined by the transformation and rendering
resolution. In particular, magnitude spectrograms do not
explicitly retain phase information. Raw IQ recordings
preserve both sampled amplitude and phase
information~\citep{o2016radio,o2018over,davaslioglu2022self},
motivating the development of signal-to-language mappings
directly from IQ inputs.

Comprehensive EM understanding requires models to process long
recordings and generate detailed descriptions of their content.
Several waveform-based approaches primarily process short
signal segments and focus on classification or concise
responses~\citep{chen2025radiollm,Shen_2026_CVPR,rostami2026largelanguagemodelsperform}. Although image-based approaches also support text generation, describing long raw-IQ recordings requires identifying multiple signal components, associating attributes with each component, and characterizing their temporal and spectral relationships. LEMMO addresses these requirements by combining long-recording processing with instruction-conditioned signal retrieval.
Its structured queries and auxiliary factual supervision
organize the retrieved evidence into scene-level properties,
object attributes, and cross-object relationships for
description generation.

\begin{figure}[!t]
\centering
\includegraphics[width=1\linewidth]{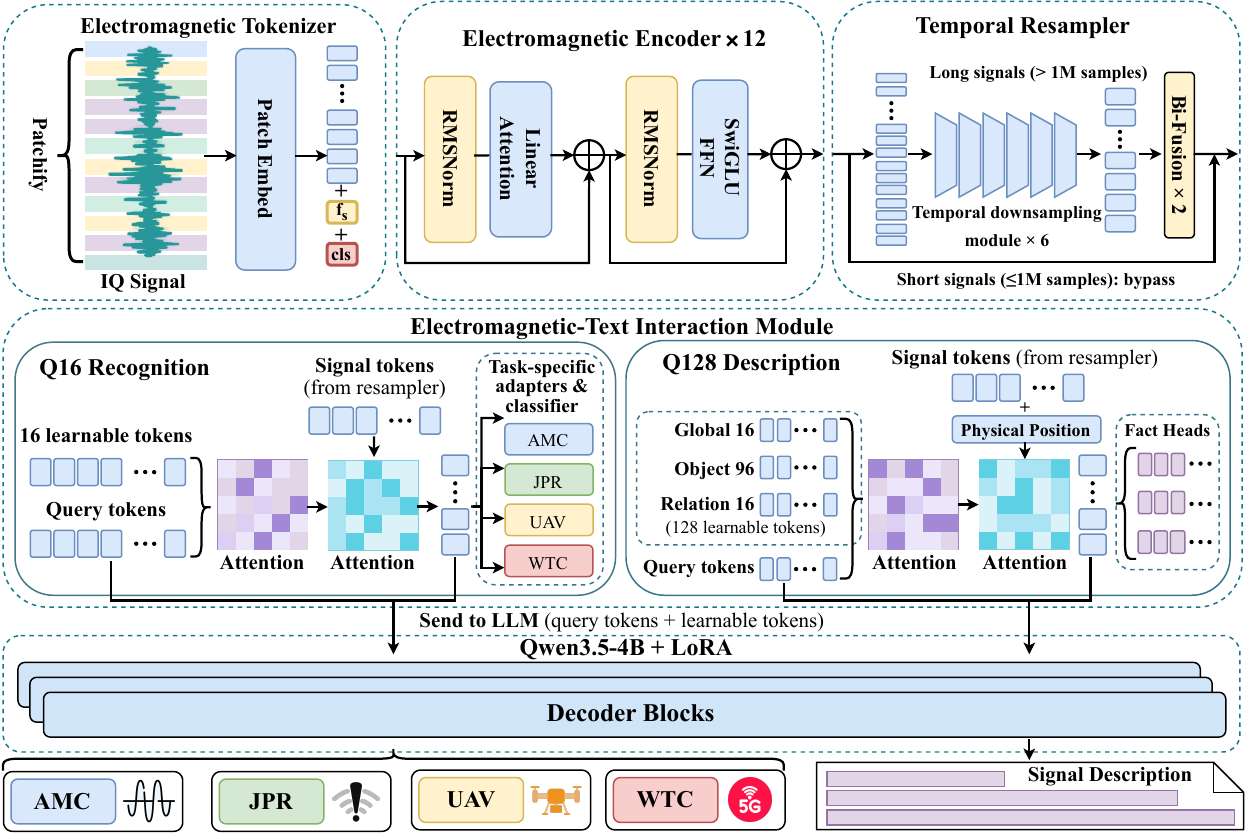}
\caption{Overview of LEMMO. Raw IQ signals are encoded and temporally
compressed for long recordings. Instruction-conditioned queries
retrieve signal evidence for a LoRA-adapted LLM, supporting
recognition with 16 queries and detailed description with
128 structured queries.}
\label{overview}
\end{figure}

\section{Method}
\label{sec:method}

\subsection{Overview}
\label{sec:overview}
LEMMO maps raw EM waveforms to natural-language responses through an instruction-conditioned signal interface. Given an IQ recording, its sampling rate, and a natural-language instruction, the model retrieves relevant signal evidence and projects it into the embedding space of a language model. As shown in Figure~\ref{overview}, LEMMO comprises an EM tokenizer, an EM encoder, a temporal resampler, an EM-text interaction module, and a Qwen3.5-4B language backbone. The tokenizer and encoder extract patch-level representations directly from IQ samples. For long recordings, the temporal resampler combines local compression with global information exchange. The interaction module then conditions learnable queries on the textual instruction before retrieving signal evidence, making the signal-to-language mapping dependent on the requested task.

\subsection{EM Tokenizer and Encoder}
\label{sec:signal_encoder}

\paragraph{Raw IQ tokenization.}
The EM tokenizer jointly normalizes the I and Q channels using a common peak-amplitude scale, placing recordings from different wireless technologies on the same scale while preserving relative amplitudes within each recording. The normalized sequence is divided into non-overlapping patches of 16 IQ samples, each mapped to a signal token and augmented with a sinusoidal positional encoding to represent patch order. A sampling-rate token and a learnable CLS token are added to the token sequence to encode the acquisition rate and provide a dedicated representation for global information, respectively.

\paragraph{Efficient waveform encoding.}
The EM encoder comprises 12 causal linear-attention blocks, with computational cost approximately linear in token-sequence length to support efficient encoding of high-rate IQ signals. We pretrain the encoder using masked waveform reconstruction, in which randomly selected signal patches are masked and reconstructed from the unmasked context. This training encourages the encoder to learn waveform structure before joint EM-language training.

\subsection{Temporal Resampler}
\label{sec:long_resampling}

Even with efficient encoding, long IQ recordings produce too many tokens for instruction-conditioned retrieval. The temporal resampler addresses this bottleneck by compressing the encoded sequence and integrating context across subsequences.

For recordings containing at most one million IQ samples, the temporal resampler is bypassed, and the EM encoder outputs are passed directly to the EM-text interaction module. For longer recordings, the encoded sequence is partitioned into consecutive subsequences, each corresponding to one million IQ samples, with a shorter final subsequence for any remaining samples. Each encoded subsequence is independently processed by a six-stage temporal downsampling module, reducing its token length by a factor of 64. The compressed subsequences are concatenated in their original order and processed by two bidirectional linear-attention layers to integrate context across subsequences. The resulting features are then passed to the EM-text interaction module.

\subsection{EM-Text Interaction Module}
\label{sec:em_text_interaction}

Different tasks require different evidence from the same
recording. The EM-text interaction module selects this evidence
through two successive cross-attention operations: learnable queries
first attend to the task instruction and then retrieve signal features. The resulting representations provide a compact,
instruction-conditioned input to the LLM. This module operates in two configurations according to the tasks: Q16 Recognition uses 16 queries for short-answer recognition tasks, whereas Q128 Description uses 128 structured queries for detailed signal descriptions.

Let $T$ denote the textual instruction-token embeddings obtained from the LLM's embedding layer, and let $H$ denote the signal features output by the Temporal Resampler, or directly by the EM encoder when resampling is bypassed. We write $\mathrm{CrossAttn}(Q,K,V)$ for cross-attention with queries $Q$, keys $K$, and values $V$.

\paragraph{Q16 Recognition.}Q16 Recognition maintains a bank of 16 learnable query tokens $Q_0$,
randomly initialized before training. These queries first attend
to the instruction embeddings:
\begin{equation}
Q_{\mathrm{inst}}
=
Q_0
+
\mathrm{CrossAttn}(Q_0,T,T).
\end{equation}
The instruction-conditioned queries then retrieve evidence from
the signal features:
\begin{equation}
Q_s
=
Q_{\mathrm{inst}}
+
\mathrm{CrossAttn}(Q_{\mathrm{inst}},H,H).
\end{equation}
This allows the instruction to guide signal retrieval
before passing to the LLM.

The retrieved queries support both answer generation and auxiliary
classification. For generation, $Q_s$ is projected into the LLM's
embedding space and supplied together with the instruction.
For classification, four task-specific residual adapters specialize
$Q_s$ for AMC, JPR, UAV recognition, and
wireless-technology classification (WTC).
Each adapter consists of layer normalization, a linear layer,
and a GELU activation within a residual connection:
\begin{equation}
Q^{t}
=
Q_s
+
\mathrm{GELU}
\left(
\mathrm{Linear}^{t}
\left(
\mathrm{LN}^{t}(Q_s)
\right)
\right),
\end{equation}
where $t \in \{\mathrm{AMC},\mathrm{JPR},\mathrm{UAV},\mathrm{WTC}\}$.
The adapted representations are passed to the corresponding
task-specific classifiers. This auxiliary path encourages the
retrieved queries to retain category-discriminative information. We train the whole Q16 module with the following loss:
\begin{equation}
\mathcal{L}_{\mathrm{Q16}}
=
\mathcal{L}_{\mathrm{LM}}
+
\lambda_{\mathrm{cls}}\mathcal{L}_{\mathrm{cls}},
\end{equation}
where $\mathcal{L}_{\mathrm{LM}}$ is the autoregressive
cross-entropy loss on reference answer tokens,
$\mathcal{L}_{\mathrm{cls}}$ is the cross-entropy loss between
the relevant task-specific classifier's predicted distribution
and the ground-truth class label, and $\lambda_{\mathrm{cls}}$
controls the auxiliary loss weight.

\paragraph{Q128 Description.}
Q128 Description uses 128 learnable queries organized into
16 global queries, 96 object queries, and 16 relation queries.
Each signal segment is treated as a signal object.
Global queries represent scene-level properties and object counts;
object queries form 16 slots with six queries per object;
and relation queries capture cross-object context.

The queries follow the same instruction-conditioning and signal-retrieval procedure as Q16 Recognition. To support temporal interpretation, we augment each signal feature $h_i$ with positional information
before retrieval:
\begin{equation}
\tilde{h}_i
=
h_i+\mathrm{PosEnc}(\rho_i),
\end{equation}
where $\rho_i$ contains the token's position in the full recording,
its physical time derived from the sampling rate, and its subsequence
index. The resulting sequence $\tilde{H}$ replaces $H$ in the
signal-retrieval cross-attention.

The retrieved queries are projected into the LLM's embedding space
for description generation. Auxiliary fact heads supervise the same
representations by predicting scene-level properties, object counts,
object-to-signal associations, signal attributes, and temporal or
spectral relations. These heads are used only during training,
and their predictions are not supplied to the LLM. The training loss of this module is
\begin{equation}
\mathcal{L}_{\mathrm{Q128}}
=
\mathcal{L}_{\mathrm{LM}}
+
\sum_{r\in\mathcal{A}(a)}
\lambda_r\mathcal{L}_{\mathrm{fact}}^{(r)},
\end{equation}
where $\mathcal{L}_{\mathrm{LM}}$ is the autoregressive
cross-entropy loss on reference description tokens and
$\mathcal{A}(a)$ denotes the fact types annotated in reference
description $a$. Each auxiliary loss uses cross-entropy for
categorical targets or smooth L1 loss for numerical targets,
weighted by $\lambda_r$. We note that unannotated fact types incur no
auxiliary loss during training.

\subsection{Supervised Fine-Tuning of LLM}
\label{sec:qwen_lora_sft}

The signal representations $Q_s$ produced by the EM-text interaction
module are projected into the embedding space of Qwen3.5-4B and
combined with the instruction embeddings $T$ to form the LLM input.
The LLM then generates a response conditioned on the instruction
and the retrieved signal evidence. We adapt Qwen3.5-4B to
EM tasks using rank-16 LoRA by minimizing the
autoregressive cross-entropy loss on reference response tokens.
The base LLM parameters remain frozen, and only its LoRA
parameters are updated.

\section{Experiments}
\label{sec:experiments}

\subsection{Datasets and Experimental Setup}
\label{sec:experimental_setup}

\paragraph{Datasets.}
Table~\ref{tab:dataset_summary} summarizes the four datasets used
in our experiments. We construct LEMMO-ShortQA and LEMMO-LongDesc
to support EM understanding tasks ranging from signal recognition
to detailed descriptions of long IQ recordings~\citep{tekbiyik2020hisarmod,oshea2016convolutional,o2016radio,schmidt2019owl,luo2025emind,swinney2021gnss,shi2025rfuav,subray2023wireless,fontaine2020subghz,fontaine2019technology,liu2017rssi,yu2024dronerfa,ren2025dronerfbdir}.
LEMMO-ShortQA provides short-answer supervision for four tasks:
AMR, JPR, UAV 
recognition, and WTC. It includs
45 classes, with input lengths ranging from 128 to 5,000,000
IQ samples. LEMMO-LongDesc supports detailed signal description,
pairing long IQ recordings with natural-language instructions
and reference descriptions. It includes 4G LTE, 5G NR,
basic-modulation, ISM, and UAV signals, with input lengths
ranging from approximately 400,000 to 8,000,000 IQ samples.

We also use EM-134K and EM-Bench~\citep{Shen_2026_CVPR}.
EM-134K provides 134,107 public signal--language pairs for
instruction tuning. EM-Bench contains 14 tasks with 300 examples
per task, comprising ten perception tasks and four
strategy-generation tasks. EM-Bench is used exclusively for
evaluation.

\begin{table}[htbp]
\centering
\caption{Dataset summary. Each split reports the number of IQ
recordings / instruction--response pairs. 
A dash indicates an unused split. IQ length is measured in complex
samples, with M denoting one million.}
\label{tab:dataset_summary}
\begin{tabular*}{\linewidth}{@{\extracolsep{\fill}}lcccc@{}}
\toprule
Dataset &
\shortstack{Train} &
\shortstack{Validation} &
\shortstack{Test} &
IQ length \\
\midrule
LEMMO-ShortQA &
\shortstack{3,832 / 15,328} &
\shortstack{870 / 3,480} &
\shortstack{846 / 3,384} &
128--5M \\
LEMMO-LongDesc &
\shortstack{16,103 / 165,328} &
\shortstack{856 / 8,813} &
\shortstack{652 / 802} &
0.4M--8M \\
EM-134K &
\shortstack{95,679 / 120,829} &
\shortstack{10,587 / 13,278} &
-- &
512/1,024 \\
EM-Bench &
-- &
-- &
\shortstack{3,370 / 4,200} &
512/1,024 \\
\bottomrule
\end{tabular*}
\end{table}

\paragraph{Baselines and evaluation.}
We compare LEMMO with MERLIN-LoRA and four general-purpose LLMs
on EM-Bench, LEMMO-ShortQA, and LEMMO-LongDesc. Note that the released MERLIN checkpoint is obtained by fully fine-tuning
the Qwen3-4B-Instruct-2507 backbone. Although this adaptation
improves EM-task performance, it is accompanied by
a substantial decline in general knowledge performance:
MMLU accuracy decreases from 57.41\% for the original backbone
to 30.00\% for MERLIN. To support EM adaptation
while preserving general knowledge, we construct and train
MERLIN-LoRA as a baseline. MERLIN-LoRA freezes the
Qwen3-4B-Instruct-2507 backbone and applies rank-16 LoRA
for language-model adaptation. In addition to evaluating
EM tasks, we assess general knowledge retention
on MMLU by comparing MERLIN and MERLIN-LoRA with their
Qwen3-4B-Instruct-2507 backbone, and LEMMO with its
Qwen3.5-4B backbone.

\subsection{EM-Bench Performance and General Knowledge Retention}
\label{sec:embench_results}

\begin{table*}[htbp]
\centering
\caption{Performance on the 14 EM-Bench tasks. Perception tasks
report accuracy (Acc., \%), and strategy-generation tasks report
ROUGE-L F1 (R-L, scaled to 0--100). Averages are computed across
tasks within each group. The highest score in each row is bold.}
\label{tab:embench_results}
\setlength{\tabcolsep}{4pt}
\begin{tabular}{@{}llrrrrrr@{}}
\toprule
Task & Metric & \textbf{LEMMO} & \shortstack{MERLIN\\-LoRA} & \shortstack{GPT-6\\Luna} & \shortstack{DeepSeek\\V4.1-Flash} & \shortstack{Gemini-3.5\\Flash-Lite} & \shortstack{Qwen3.8\\Flash} \\
\midrule
\multicolumn{8}{@{}l}{\emph{Perception}} \\
CJR & Acc. & \textbf{83.67} & 72.33 & 13.00 & 11.67 & 15.33 & 21.33 \\
MOD & Acc. & \textbf{36.00} & 27.00 & 0.00 & 1.33 & 9.33 & 10.33 \\
Radar bandwidth & Acc. & \textbf{58.33} & 27.00 & 0.00 & 0.00 & 0.33 & 1.33 \\
Radar duty cycle & Acc. & \textbf{54.33} & 30.67 & 1.67 & 0.67 & 14.67 & 11.00 \\
Radar NP & Acc. & \textbf{82.00} & 81.00 & 9.33 & 1.00 & 23.00 & 18.33 \\
Radar PRF & Acc. & \textbf{41.33} & 26.33 & 0.00 & 0.00 & 4.00 & 3.00 \\
Radar pulse width & Acc. & \textbf{37.33} & 30.33 & 0.00 & 0.00 & 7.00 & 3.33 \\
PI & Acc. & \textbf{79.67} & 71.00 & 3.33 & 10.00 & 7.67 & 16.00 \\
RJR & Acc. & \textbf{73.67} & 45.67 & 24.67 & 12.67 & 19.33 & 27.67 \\
SD & Acc. & 46.67 & 24.00 & 27.00 & 36.33 & \textbf{56.67} & 56.00 \\
\addlinespace[2pt]
\textbf{Perception average} & Acc. & \textbf{59.30} & 43.53 & 7.90 & 7.37 & 15.73 & 16.83 \\
\midrule
\multicolumn{8}{@{}l}{\emph{Generation}} \\
Anti-CJ & R-L & 50.68 & \textbf{51.14} & 0.22 & 1.31 & 7.81 & 8.27 \\
Anti-RJ & R-L & \textbf{43.85} & 35.12 & 0.64 & 1.30 & 7.64 & 9.16 \\
CJS & R-L & 39.84 & \textbf{44.64} & 1.32 & 3.77 & 8.36 & 7.04 \\
RJS & R-L & \textbf{56.17} & 42.14 & 1.35 & 5.60 & 10.28 & 9.37 \\
\addlinespace[2pt]
\textbf{Generation average} & R-L & \textbf{47.64} & 43.26 & 0.88 & 3.00 & 8.52 & 8.46 \\
\bottomrule
\end{tabular}%
\end{table*}

Table~\ref{tab:embench_results} shows that LEMMO achieves
the highest scores on 11 of the 14 EM-Bench tasks.
It outperforms MERLIN-LoRA on all ten perception tasks,
achieving 59.30\% mean accuracy (+15.77 percentage points)
and ranking first on nine tasks.
For strategy generation, LEMMO leads on Anti-RJ and RJS
and achieves the highest mean ROUGE-L F1 of 47.64,
exceeding MERLIN-LoRA by 4.38 points.
General-purpose LLMs perform substantially worse on most
tasks, although Gemini-3.5-Flash-Lite leads on SD.
These results demonstrate LEMMO's effectiveness in both
EM perception and strategy generation.

\begin{table*}[htbp]
\centering
\caption{MMLU accuracy (\%) for EM-adapted models and their
original language backbones. MERLIN and MERLIN-LoRA use
Qwen3-4B-Instruct-2507; LEMMO uses Qwen3.5-4B.
The highest score in each column is bold.}
\label{tab:mmlu_results}
\setlength{\tabcolsep}{9pt}
\begin{tabular}{lccccc}
\toprule
Model & Overall & STEM & Humanities & Social Sciences & Other \\
\midrule
MERLIN                     & 30.00 & 28.45 & 27.97 & 32.95 & 31.70 \\
MERLIN-LoRA                & 57.55 & 53.60 & 48.33 & 70.26 & 62.92 \\
Qwen3-4B-Instruct-2507     & 57.41 & 50.02 & 53.41 & 68.38 & 60.12 \\
Qwen3.5-4B                 & 64.34 & 58.77 & \textbf{59.70} & 71.86 & 69.58 \\
\midrule
\textbf{LEMMO}                      & \textbf{64.98} & \textbf{60.04} &
59.49 & \textbf{72.64} & \textbf{70.71} \\
\bottomrule
\end{tabular}
\end{table*}

Table~\ref{tab:mmlu_results} reports MMLU accuracy before
and after EM adaptation. LEMMO achieves 64.98\% overall accuracy, compared with 64.34\% for its Qwen3.5-4B backbone. MERLIN-LoRA similarly preserves its backbone's accuracy (57.55\% versus 57.41\%), whereas the released MERLIN checkpoint scores 30.00\%, a decline of 27.41 percentage points. These findings show that LEMMO achieves strong EM-task performance while preserving general knowledge.

\subsection{EM Recognition on LEMMO-ShortQA}
\label{sec:shortqa_results}

Table~\ref{tab:shortqa_results} reports open-ended recognition
accuracy on LEMMO-ShortQA. LEMMO achieves the highest accuracy
on all four tasks, with a mean of 70.05\%, exceeding
MERLIN-LoRA by 14.31\%.
It improves AMC, JPR, and WTC by 9.71, 7.77, and 7.89
\%, respectively. The largest gain occurs in UAV
recognition, where accuracy increases from 41.88\% to
73.75\% (+31.87 \%). All four general-purpose LLMs
obtain substantially lower accuracy across the tasks.

\begin{table*}[htbp]
\centering
\caption{Recognition accuracy (\%) on the LEMMO-ShortQA test set.
Average denotes the unweighted mean across the four tasks.
The highest score in each column is bold.}
\label{tab:shortqa_results}
\begin{tabular*}{\textwidth}{@{\extracolsep{\fill}}lrrrrr@{}}
\toprule
Model & AMC & JPR & UAV & WTC & Average \\
\midrule
GPT-6 Luna & 0.32 & 4.00 & 0.00 & 0.08 & 1.10 \\
DeepSeek-V4.1-Flash & 0.81 & 3.12 & 0.00 & 1.95 & 1.47 \\
Gemini-3.5-Flash-Lite & 2.35 & 10.29 & 0.00 & 2.42 & 3.77 \\
Qwen3.8-Flash & 4.14 & 3.48 & 0.00 & 4.14 & 2.94 \\
MERLIN-LoRA & 60.01 & 61.47 & 41.88 & 59.61 & 55.74 \\
\midrule
\textbf{LEMMO} & \textbf{69.72} & \textbf{69.24} & \textbf{73.75} &
\textbf{67.50} & \textbf{70.05} \\
\bottomrule
\end{tabular*}
\end{table*}

Table~\ref{tab:shortqa_iq_length} summarizes the input lengths
for each task. UAV recognition uses the longest
recordings, each containing five million IQ samples, and
also exhibits the largest improvement over MERLIN-LoRA.
Together with the gains on shorter inputs, this result
demonstrates LEMMO's effectiveness across recognition tasks
with widely varying signal lengths.

\begin{table}[htbp]
\centering
\caption{Input IQ lengths for each LEMMO-ShortQA task,
measured in complex samples.}
\label{tab:shortqa_iq_length}
\begin{tabular*}{1\linewidth}{@{\extracolsep{\fill}}cccc@{}}
\toprule
AMC & JPR & UAV & WTC \\
\midrule
128--1,024 samples & 128--16,368 samples & 5,000,000 samples & 4,096 samples \\
\bottomrule
\end{tabular*}
\end{table}


\subsection{Long-Recording Description on LEMMO-LongDesc}

We evaluate two tasks on the LEMMO-LongDesc test set:
multi-signal scene overview and per-signal detailed description.
Table~\ref{tab:longdesc_multisignal_two_types} reports both
text-overlap and fact-level metrics. ROUGE-1 F1 (R-1), ROUGE-2 F1 (R-2), ROUGE-L F1 (R-L), BLEU-4, and Token F1 measure agreement with reference descriptions. Category F1 evaluates the recovery of categorical facts, such as modulation type, while Value F1 evaluates numerical facts, including signal counts, start and end times, and center frequencies.

\begin{table*}[htbp]
\centering
\caption{Description performance on the LEMMO-LongDesc test set.
All scores are scaled to 0--100. The highest score for each
metric within each task is bold.}
\label{tab:longdesc_multisignal_two_types}
\setlength{\tabcolsep}{2pt}
\begin{tabular*}{\textwidth}{@{}l@{\extracolsep{\fill}}rrrrrrr@{}}
\toprule
& \multicolumn{5}{c}{Text Metrics} & \multicolumn{2}{c}{Fact Metrics} \\
\cmidrule(lr){2-6}\cmidrule(l){7-8}
Model & R-1 & R-2 & R-L & BLEU-4 & Token F1 & Category F1 & Value F1 \\
\midrule
\multicolumn{8}{@{}l}{\emph{Multi-signal scene overview task}} \\
GPT-6 Luna & 22.79 & 6.26 & 11.49 & 3.86 & 23.51 & 0.00 & 0.00 \\
DeepSeek-V4.1-Flash & 21.88 & 5.95 & 10.36 & 2.64 & 23.01 & 0.00 & 0.50 \\
Gemini-3.5-Flash-Lite & 20.32 & 5.33 & 11.25 & 0.93 & 21.89 & 0.10 & 0.00 \\
Qwen3.8-Flash & 20.62 & 5.50 & 10.69 & 1.99 & 22.02 & 1.20 & 0.30 \\
MERLIN-LoRA & 72.76 & 61.37 & 66.17 & 49.43 & 70.31 & 39.76 & 36.10 \\
\cmidrule(lr){1-8}
\textbf{LEMMO} & \textbf{76.45} & \textbf{65.43} & \textbf{70.83} & \textbf{50.94} & \textbf{74.80} & \textbf{44.83} & \textbf{44.83} \\
\midrule
\multicolumn{8}{@{}l}{\emph{Per-signal detailed description task}} \\
GPT-6 Luna & 31.78 & 8.62 & 13.83 & 4.79 & 32.02 & 0.00 & 0.48 \\
DeepSeek-V4.1-Flash & 28.22 & 8.21 & 12.53 & 3.60 & 28.94 & 0.27 & 2.86 \\
Gemini-3.5-Flash-Lite & 26.48 & 7.52 & 14.47 & 2.97 & 27.51 & 0.00 & 0.00 \\
Qwen3.8-Flash & 26.05 & 7.53 & 11.81 & 2.98 & 27.16 & 0.75 & 0.03 \\
MERLIN-LoRA & 68.41 & 54.16 & 64.71 & \textbf{43.63} & 66.83 & 34.43 & 19.61 \\
\cmidrule(lr){1-8}
\textbf{LEMMO} & \textbf{68.51} & \textbf{54.82} & \textbf{65.11} & 42.96 & \textbf{67.77} & \textbf{41.42} & \textbf{24.06} \\
\bottomrule
\end{tabular*}
\end{table*}

LEMMO leads on all seven metrics for multi-signal scene overview.
Compared with MERLIN-LoRA, it improves ROUGE-L F1 by 4.66 points
and Token F1 by 4.49 points. The gains are larger for factual
content: Category F1 increases from 39.76 to 44.83 (+5.07 points),
and Value F1 increases from 36.10 to 44.83 (+8.73 points). For per-signal detailed description, LEMMO achieves the highest
scores on six of seven metrics. Its main improvements are in
Category F1, which rises from 34.43 to 41.42 (+6.99 points),
and Value F1, which rises from 19.61 to 24.06 (+4.45 points).
Text-overlap gains are smaller, while BLEU-4 decreases by
0.67 points. General-purpose LLMs obtain substantially lower
scores on both tasks, particularly on the fact-level metrics.
These results demonstrate LEMMO's effectiveness in describing
long recordings, with consistent improvements in the recovery
of categorical and numerical signal facts.

\subsection{Effect of Input Signal Length}
\label{sec:ablations}

Table~\ref{tab:input_length} evaluates factual prediction
using the same instructions with input lengths ranging
from 4K to 4M IQ samples. Category F1 increases consistently
with input length, rising from 42.13 to 68.45.
Value F1 also peaks at 4M samples, increasing from 13.03
to 35.24, although it decreases at 1M samples.
Relative to 4K inputs, 4M inputs improve Category F1
and Value F1 by 26.32 and 22.21 points, respectively.
These results indicate that LEMMO can use longer signal
observations to improve factual prediction, highlighting 
the effectiveness of long-recording EM modeling.

\begin{table}[t]
\centering
\caption{Effect of input IQ length on fact metrics. The same questions are
evaluated at each length.}
\label{tab:input_length}
\begin{tabular*}{\linewidth}{@{\extracolsep{\fill}}lccccc c@{}}
\toprule
Input IQ length & 4K & 64K & 256K & 1M & 4M & 4K$\rightarrow$4M \\
\midrule
Category F1 & 42.13 & 49.31 & 61.69 & 63.08 & \textbf{68.45} & \textbf{+26.32} \\
Value F1    & 13.03 & 24.76 & 24.77 & 20.00 & \textbf{35.24} & \textbf{+22.21} \\
\bottomrule
\end{tabular*}
\end{table}

\section{Conclusion}

We introduced LEMMO, a framework for instruction-driven EM
understanding from raw IQ recordings of up to eight million
samples. LEMMO combines hierarchical long-sequence processing
with an EM--text interface that conditions learnable queries
on instructions before retrieving signal evidence. Its
complementary query designs support both recognition and
detailed description: compact queries receive discriminative
supervision, while structured queries organize scene, object,
and relation information with auxiliary factual supervision.
We also constructed LEMMO-ShortQA and LEMMO-LongDesc to support
training and evaluation across these tasks. LEMMO significantly outperforms MERLIN-LoRA on average across EM-Bench, on all four
LEMMO-ShortQA tasks, and on most LEMMO-LongDesc metrics,
while retaining its LLM backbone's MMLU performance.
Input-length experiments further demonstrate improved factual
prediction with longer observations, highlighting the value
of combining long raw-IQ processing with instruction-conditioned,
structured evidence retrieval for EM understanding.

\bibliography{main}
\bibliographystyle{plainnat}

\appendix

\section{Detailed Model Structure}
\label{app:model}

This section provides further details on the model components introduced in Section~3. The model divides IQ samples into patches and encodes them. It compresses the resulting features when the recording exceeds one million samples, then passes signal information to Qwen3.5-4B. Q16 Recognition and Q128 Description share the same tokenizer and encoder architecture but use different queries and training objectives.

\subsection{EM Tokenizer and EM Encoder}

\paragraph{IQ normalization and patching.}
Let $X\in\mathbb{R}^{L\times 2}$ be an IQ recording with $L$ complex samples. Its two columns contain the in-phase (I) and quadrature (Q) components. Both channels are normalized by the same record-level scale,  $s(X)=\max_{t,c}|X_{t,c}|$, the maximum absolute value across all I and Q values in the recording.

After normalization, every 16 consecutive complex samples form one non-overlapping patch. Each patch contains 32 real values, which a shared linear layer maps to a 768-dimensional vector. A recording of length $L$ yields $\lfloor L/16\rfloor$ complete patches; a final segment shorter than 16 samples is omitted. Each independently encoded sequence also receives a learnable CLS token and a sampling-rate token. The sampling-rate token is produced by a learnable multilayer perceptron (MLP) from fixed-frequency features of the sampling rate. Fixed sinusoidal position encoding restarts at the beginning of each sequence.

Multiple short recordings are placed in one computation container with a total capacity of one million IQ samples. The encoder receives the boundaries of each recording and computes the attention separately within each recording, preventing information exchange between recordings. A recording longer than one million samples is divided into consecutive regions. All regions use the same patch embedding layer and EM Encoder.

The EM Encoder has 12 layers. Each layer has a hidden size of 768, 12 attention heads, and a feed-forward hidden size of 3072. The layers use RMSNorm, FLA causal LinearAttention, and a SwiGLU feed-forward network, with residual connections after the attention and feed-forward sublayers. The patch embedding and encoder parameters are shared across input lengths.

Recordings of at most one million complex IQ samples bypass the Temporal Resampler. Longer recordings are divided into regions of at most one million samples. For each complete region, 62,500 encoded patch tokens are reduced to 977 tokens by six temporal downsampling stages and then pooled to 800 tokens, as detailed in Appendix A.2. 

\paragraph{Masked waveform reconstruction.}
During pretraining, a random subset of patches is masked, and the encoder processes only visible patches. A shared reconstruction decoder with eight layers and a hidden size of 384 inserts learnable mask tokens at the missing positions and predicts the 32 real values of each masked patch.

Before computing the loss, the 32 target values of each masked patch $p$ are standardized using the mean $\mu_p$ and variance $\sigma_p^2$. The reconstruction loss averages squared error over the 32 values and all masked patches:
\begin{equation}
\mathcal{L}_{\mathrm{rec}}=
\frac{1}{|\mathcal{M}|}\sum_{p\in\mathcal{M}}\frac{1}{32}
\left\|\hat{x}_p-\frac{x_p-\mu_p}{\sqrt{\sigma_p^2+\epsilon}}\right\|_2^2,
\qquad \epsilon=10^{-6}.
\label{eq:rec}
\end{equation}
Here $\mathcal{M}$ is the set of masked patches, $x_p$ is the original patch, and $\hat{x}_p$ is the decoder prediction. The decoder is used for pretraining and is discarded for question answering.

\subsection{Temporal Resampler}

The Temporal resampler processes only recordings longer than one million IQ samples. The recording is divided in temporal order into regions of at most one million samples. The EM Encoder encodes each region independently, and the resampler then compresses its encoded features.

The Temporal Resampler first applies six temporal downsampling stages, with an overall downsampling factor of 64. Each stage contains a low-pass downsampling path and a gated learnable convolutional path. The low-pass path reduces aliasing, while the convolutional path captures local variation. For a complete region containing one million complex IQ samples, the encoder produces 62,500 patch tokens. The six-stage downsampling stack reduces this sequence to 977 tokens, after which an average-pooling layer reduces it to 800 tokens. A shorter final region receives an output budget proportional to its number of valid patch tokens. 

The compressed features are concatenated in the original temporal order. The two bidirectional linear-attention layers then exchange information between regions. Thus, the encoder represents the signal within each region before compression, and information is exchanged between regions after compression.

The compressed features retain their temporal order. At the maximum supported length of eight million samples, eight complete regions yield 6,400 features for the EM-Text Interaction Mechanism.

\subsection{EM-Text Interaction Mechanism}

This module uses learnable query vectors to extract information from the question and signal for answer generation. The queries first attend to the question tokens and then to the signal features, following Eqs. (1)–(2) for Q16 Recognition and Eqs. (4)–(6) for Q128 Description. Q16 Recognition uses 16 queries, whereas Q128 Description uses 128. Each query output is projected from 768 dimensions to the 2560-dimensional embedding space of Qwen3.5-4B and enters the language model as a separate signal token.

\paragraph{Q16 Recognition.}
This configuration uses 16 learnable query vectors for AMC, JPR, UAV recognition, and WTC. The queries first attend to the question tokens and then to the signal features, producing the retrieved representations $Q_s$. These representations are projected into the Qwen3.5-4B embedding space and provided to the LLM together with the question tokens for answer generation.

During supervised training, $Q_s$ also enters an auxiliary classification branch. Each task has a residual adapter consisting of LayerNorm, a 768 → 64 linear layer, GELU, and a 64 → 768 linear layer, followed by a residual addition. Learned attention weights aggregate the adapted representations into a task representation for the corresponding classifier. The adapters and classifiers provide auxiliary classification supervision during training. At inference, answer generation uses the direct $Q_s$-to-LLM path without invoking the classification branch.

\paragraph{Q128 Description.}
This instance uses 128 queries: 16 global queries for information about the full recording, 96 object-feature queries for individual signal objects, and 16 relation queries for interactions between objects. The 96 object-feature queries form 16 groups of six, with one group per object. Query-group, object-index, and feature-type embeddings specify the purpose of each query before it attends to the question.

Q128 also adds three position coordinates to each signal feature: its relative position in the full recording, normalized log physical time computed from the sampling rate, and its position within the long-record region structure. Each coordinate retains its original value and receives sine and cosine features at eight Fourier frequencies, giving $3\times(1+8\times2)=51$ dimensions in total. A $51\to256\to768$ projection maps this position vector to the signal-feature dimension. The projected vector is added to the signal feature and followed by LayerNorm. Position information is added before the queries attend to the signal. The resulting 128 query representations support both language generation and auxiliary fact prediction.

Four types of heads provide auxiliary predictions. Scene-level heads predict global categories and counts. Object-level heads predict existence, categories, numeric attributes, and multi-label attributes. A binding head scores associations between object representations and time-ordered signal features. Pairwise heads predict relations between object representations; these heads are separate from the 16 relation queries. Their predictions are used only for training and are not converted into fact tokens for Qwen3.5-4B.

For answer $a$, let $\mathcal{A}(a)$ be the set of annotated fact types. The training loss combines language-model loss with auxiliary losses for those fact types:
\begin{equation}
\mathcal{L}_{\mathrm{desc}}=\mathcal{L}_{\mathrm{LM}}+
\sum_{r\in\mathcal{A}(a)}\lambda_r\mathcal{L}_{\mathrm{fact}}^{(r)}.
\label{eq:desc}
\end{equation}
Here $\lambda_r$ is the weight of fact type $r$. Unannotated fact types do not contribute to the auxiliary loss and are not treated as negative examples.

\section{Training and Implementation Details}
\label{app:training}

This section describes three experimental configurations: Q16 Recognition for LEMMO-ShortQA, Q128 Description for LEMMO-LongDesc, and the direct encoder-to-LLM interface for EM-Bench. The configurations use the same EM Encoder architecture with configuration-specific weights and training stages. All language-generation objectives use causal cross-entropy computed on answer tokens. 

\subsection{Signal Representation Pretraining}
\label{app:pretraining}

The EM Encoder is pretrained by masked waveform reconstruction
on AMC, JPR, UAV, and WTC IQ recordings, sampled
in proportions of 20\%, 30\%, 30\%, and 20\%, respectively.
Both I and Q channels are normalized by the maximum absolute value across
the two channels of each recording. The EM Tokenizer maps
each patch to a 768-dimensional token. The patch embedding and
reconstruction decoder are shared across all recording lengths.

With a 50\% mask ratio, the EM Encoder processes visible
patch tokens, and an eight-block decoder with hidden width 384
reconstructs the masked patches. Each target patch is standardized using
the mean and variance of its 32 real-valued elements. The loss is the
mean squared reconstruction error over masked patches.

Pretraining runs for three passes on four H100 GPUs. AdamW uses a peak
learning rate of $5\times10^{-5}$, weight decay 0.01, a 5\% warmup,
and gradient clipping at norm 1.0. For Q128 Description, the EM
Encoder undergoes two further passes of masked reconstruction on 4G LTE,
5G NR, basic-modulation, ISM, and UAV recordings. This stage uses the same
EM Tokenizer with patch size 16 and a 50\% mask ratio. The peak
learning rates are $10^{-5}$ and $10^{-6}$ in sequence.

\subsection{Training Q16 Recognition}
\label{app:training_recognition}

Q16 Recognition is trained on LEMMO-ShortQA, which contains 15,328 training
question--answer pairs from 3,832 signals and 3,480 validation pairs from
870 signals. Its four tasks are AMC, JPR, UAV recognition, and WTC.
Each signal has four question templates. Training samples are balanced
across tasks, source datasets, and labels.

The Temporal Resampler is first trained to compress EM Encoder
features. For recordings longer than one million complex IQ samples, the Temporal Resampler reduces the 62,500 patch tokens of each complete one-million-sample region to 977 tokens through six downsampling stages, followed by average pooling to obtain 800 tokens. Two bidirectional fusion layers integrate context across the compressed regions.  Training runs for three passes with
a peak learning rate of $3\times10^{-4}$. The objective combines
compressed-feature, cosine, temporal-difference, and masked-fusion losses
with weights $1$, $0.25$, $0.25$, and $1$, respectively.

Q16 Recognition uses 16 question-conditioned queries in the
EM-Text Interaction Mechanism. Q16 Recognition is first
trained with the EM Encoder, Temporal Resampler, and
Qwen3.5-4B frozen. The final two EM Encoder blocks are then
unfrozen for further alignment with Qwen3.5-4B. Language-model adaptation
uses rank-16 LoRA with scaling parameter 32 and dropout 0.05.
The Qwen3.5-4B base parameters remain frozen. During supervised training, the retrieved query representations $Q_s$ support both answer generation and auxiliary classification. The language-generation branch projects $Q_s$ directly into the LLM embedding space. The classification branch applies task-specific residual adapters with bottleneck width 64, followed by attention pooling and classification. The final joint stage optimizes the answer-generation loss and the auxiliary classification loss: 
\begin{equation}
\mathcal{L}_{\mathrm{recog}}
=
\mathcal{L}_{\mathrm{LM}}
+
0.5\,\mathcal{L}_{\mathrm{cls}} .
\label{eq:app_recog_loss}
\end{equation}
At inference, only the language-generation branch is used. 

The final joint stage runs for three passes on four H100 GPUs. The peak
learning rates are $10^{-7}$ for the EM Encoder,
$2\times10^{-6}$ for Q16 Recognition, $5\times10^{-5}$ for the
task-specific adapters and classifiers, and $5\times10^{-6}$ for
Qwen3.5-4B LoRA. AdamW uses $(\beta_1,\beta_2)=(0.9,0.95)$,
weight decay 0.01, a 5\% warmup, and gradient clipping at norm 1.0.
Checkpoint selection uses label-balanced validation accuracy across
the four tasks.

\subsection{Training Q128 Description}
\label{app:training_description}

Q128 Description is trained on LEMMO-LongDesc, which contains 165,328 training
question--answer pairs from 16,103 signals and 8,813 validation pairs from
856 signals. Its five signal families are 4G LTE, 5G NR, basic modulation,
ISM, and UAV. Question--answer pairs are grouped by physical signal.
Each training pass samples five questions per physical signal.

The Temporal Resampler processes recordings containing up to eight million complex IQ samples using the downsampling and average-pooling procedure described in Appendix A.2. Each complete one-million-sample region produces 800 tokens, while a shorter final region receives an output budget proportional to its number of valid patch tokens. The Temporal Resampler is trained for three passes with
a peak learning rate of $3\times10^{-4}$ and a 25\% feature-mask ratio.
The objective combines compressed-feature, cosine, temporal-difference,
and masked-fusion losses with weights $1$, $0.25$, $0.25$, and $1$,
respectively.

Q128 Description uses 16 global queries, 96 object-feature queries, and
16 relation queries in the EM-Text Interaction Mechanism.
The object-feature queries form 16 groups of six. Initial alignment
proceeds in three stages. Q128 Description is trained first, followed by
adaptation of the final two EM Encoder blocks and then
training of rank-16 Qwen3.5-4B LoRA. Structured training uses auxiliary
fact heads to apply scene-level, object-level, binding, and pairwise-relation
supervision to the query representations. Auxiliary fact predictions
are not converted to input tokens for Qwen3.5-4B.
For answer $a$, the training objective is
\begin{equation}
\mathcal{L}_{\mathrm{desc}}
=
\mathcal{L}_{\mathrm{LM}}
+
\sum_{r\in\mathcal{A}(a)}
\lambda_r\mathcal{L}^{(r)}_{\mathrm{fact}} ,
\label{eq:app_longdesc_training_loss}
\end{equation}
where $\mathcal{A}(a)$ is the set of fact types annotated for answer $a$,
and $\lambda_r$ weights the auxiliary loss for fact type $r$.

During structured training, the final four, final eight, and all twelve
EM Encoder blocks are progressively unfrozen.
The cross-region fusion layers in the Temporal Resampler are trainable
during these stages. The final joint stage also trains Q128 Description,
the auxiliary fact heads, and rank-16 Qwen3.5-4B LoRA.
The Qwen3.5-4B base parameters remain frozen throughout training.

The final stage uses a peak learning rate of $2\times10^{-8}$ at the top
of the EM Encoder, with layer-wise learning-rate decay of
0.85. The other peak learning rates are $5\times10^{-8}$ for the
cross-region fusion layers in the Temporal Resampler, $3\times10^{-6}$
for Q128 Description, and $10^{-5}$ for Qwen3.5-4B LoRA. AdamW uses
$(\beta_1,\beta_2)=(0.9,0.95)$, weight decay 0.05, a 5\% warmup,
and gradient clipping at norm 1.0. Validation covers all
8,813 question--answer pairs. Checkpoint selection uses both validation
answer loss and the evaluation results for auxiliary fact predictions.

\subsection{EM-Bench Interface}
\label{app:public_setting}

For the EM-Bench comparison, we adopt the same direct encoder-to-LLM interface as MERLIN. The EM encoder converts the IQ recording into signal tokens, which are projected into the LLM embedding space and passed directly to Qwen3.5-4B together with the question tokens. This configuration uses a per-token projector rather than the Q16 or Q128 query-based interaction module.

We train and validate our interface on the EM-134K corpus, which is split into a training set of 120,829 question–answer pairs from 95,679 signals and a validation set of 13,278 pairs from 10,587 signals. EM-Bench is held out as an independent test set; recordings in both EM-134K and EM-Bench use 512 or 1,024 IQ samples. These lengths yield 32 or 64 time-ordered patch features,
respectively. A shared per-token projector maps each feature into the
Qwen3.5-4B embedding space:
\[
\operatorname{LN}(768)
\rightarrow
\operatorname{Linear}(768,1536)
\rightarrow
\operatorname{GELU}
\rightarrow
\operatorname{Linear}(1536,2560)
\rightarrow
\operatorname{LN}(2560).
\]
All projected features are passed directly to Qwen3.5-4B.

Training comprises five stages. The first trains the projector with the
EM Encoder and Qwen3.5-4B frozen. The next three stages
progressively unfreeze the final two, final six, and all twelve
EM Encoder blocks. The final stage jointly trains the
EM Encoder, projector, and rank-16 Qwen3.5-4B LoRA.
The Qwen3.5-4B base parameters remain frozen. Training samples are
balanced across the EM-134K question families.

The projector peak learning rates across the five stages are
$2\times10^{-4}$, $5\times10^{-5}$, $4\times10^{-5}$,
$3\times10^{-5}$, and $2\times10^{-5}$, respectively. The peak
learning rates at the top of the EM Encoder in the final
four stages are $10^{-5}$, $8\times10^{-6}$, $5\times10^{-6}$,
and $2\times10^{-6}$, with layer-wise learning-rate decay of 0.8.
The final joint stage uses a Qwen3.5-4B LoRA learning rate of
$2\times10^{-5}$. Training runs on three H100 GPUs with a 5\% warmup
and gradient clipping at norm 1.0. Checkpoint selection gives equal weight
to the ten-task perception group and the four-task generation group
on the validation set.

\section{Datasets and Evaluation Protocols}
\label{app:datasets}

Table~\ref{tab:app_datasets} summarizes the
sizes, IQ lengths, and roles of the four datasets.

\begin{table}[htbp]
\centering
\caption{Signal and QA counts used in the reported experiments.}
\label{tab:app_datasets}
{\setlength{\tabcolsep}{3pt}
\begin{tabular}{@{}lcccp{0.13\linewidth}p{0.16\linewidth}@{}}
\toprule
Dataset &
\shortstack{Train\\signals / QA} &
\shortstack{Validation\\signals / QA} &
\shortstack{Test\\signals / QA} &
IQ length &
Primary use \\
\midrule
LEMMO-ShortQA &
\shortstack{3,832 /\\15,328} &
\shortstack{870 /\\3,480} &
\shortstack{846 /\\3,384} &
128-5M &
Q16 Recoginition training and testing \\
LEMMO-LongDesc &
\shortstack{16,103 /\\165,328} &
\shortstack{856 /\\8,813} &
\shortstack{652 /\\802} &
0.4-8M &
Q128 Description training and testing \\
EM-134K &
\shortstack{95,679 /\\120,829} &
\shortstack{10,587 /\\13,278} &
-- &
512 or 1,024 &
EM-Bench training and validating \\
EM-Bench &
-- &
-- &
\shortstack{3,370 /\\4,200} &
512 or 1,024 &
EM-Bench testing \\
\bottomrule
\end{tabular}
}
\end{table}

\subsection{LEMMO-ShortQA}
\label{app:datasets_short}

LEMMO-ShortQA requires models to generate short answers from IQ signals and
natural-language questions. It covers AMC, JPR, UAV recognition, and WTC,
with 45 target labels in total. Table~\ref{tab:app_short_split} reports the
signal and QA counts for each task and split.

\begin{table}[htbp]
\centering
\caption{LEMMO-ShortQA composition. Each physical signal has four question
phrasings for the same task.}
\label{tab:app_short_split}
{\setlength{\tabcolsep}{3pt}
\begin{tabular}{@{}lcccc@{}}
\toprule
Task &
Labels &
\shortstack{Train\\signals / QA} &
\shortstack{Validation\\signals / QA} &
\shortstack{Test\\signals / QA} \\
\midrule
Modulation recognition & 13 &
\shortstack{1,856 /\\7,424} &
\shortstack{456 /\\1,824} &
\shortstack{432 /\\1,728} \\
Jamming-pattern recognition & 12 &
\shortstack{1,088 /\\4,352} &
\shortstack{182 /\\728} &
\shortstack{182 /\\728} \\
UAV recognition & 10 &
\shortstack{120 /\\480} &
\shortstack{40 /\\160} &
\shortstack{40 /\\160} \\
Wireless-technology recognition & 10 &
\shortstack{768 /\\3,072} &
\shortstack{192 /\\768} &
\shortstack{192 /\\768} \\
\midrule
Total & 45 &
\shortstack{3,832 /\\15,328} &
\shortstack{870 /\\3,480} &
\shortstack{846 /\\3,384} \\
\bottomrule
\end{tabular}
}
\end{table}

Each physical signal has four fixed question phrasings for its task.
Consequently, the QA count is four times the signal count in every split. For
example, a modulation signal may be paired with questions such as ``What
modulation scheme does this signal use?'' and ``Identify the modulation
class.''

The modulation labels are 16PAM, 16QAM, 2FSK, 4PAM, AM-DSB, AM-DSB-SC,
AM-LSB, AM-USB, CPFSK, FM, GFSK, PM, and QPSK.

The jamming-pattern labels are AM\_Jamming, Comb\_Jamming, FM\_Jamming,
Multi\_Tone, Narrowband, No\_Interference, Protocol\_Bluetooth,
Protocol\_WiFi, Pulse\_DME, Single\_Tone, Sweep\_Chirp, and Wideband.

The UAV labels are DJI Avata 2, DJI FPV Combo, DJI Mavic 3 Pro,
Devention DEVO, Fly Sky FS I6X, Herelink HX4, JR Propo XG14, SIYI MK15,
WFLY ET10, and Yunzhuo H12.

The wireless-technology labels are 5G-NR, DVB-T, IEEE 802.11ah, IEEE 802.15.4
SUN-OFDM, IEEE 802.15.4g, LTE, LoRa SF12, LoRa SF7, Sigfox, and Wi-Fi.

Table~\ref{tab:app_short_sources} lists the source datasets and the numbers of
signals assigned to the training, validation, and test splits.

\begin{table}[htbp]
\centering
\caption{Source datasets and signal counts for LEMMO-ShortQA.}
\label{tab:app_short_sources}
{\setlength{\tabcolsep}{3pt}
\begin{tabular}{@{}p{0.22\linewidth}p{0.42\linewidth}rrr@{}}
\toprule
Task & Source dataset & Train & Validation & Test \\
\midrule
Modulation recognition & HisarMod2019.1~\citep{tekbiyik2020hisarmod} & 704 & 168 & 144 \\
Modulation recognition & RML2016.04C~\citep{oshea2016convolutional} & 384 & 96 & 96 \\
Modulation recognition & RML2016.10A~\citep{o2016radio} & 384 & 96 & 96 \\
Modulation recognition & RML2016.10B~\citep{o2016radio} & 384 & 96 & 96 \\
Jamming-pattern recognition &
CRAWDAD OWL-InIT / OWL Interference Dataset~\citep{schmidt2019owl} & 128 & 32 & 32 \\
Jamming-pattern recognition & EM-Infer-Comm~\citep{luo2025emind} & 576 & 144 & 144 \\
Jamming-pattern recognition & GNSS Jamming Dataset~\citep{swinney2021gnss} & 384 & 6 & 6 \\
UAV recognition & RFUAV~\citep{shi2025rfuav} & 120 & 40 & 40 \\
Wireless-technology recognition &
Real-world Wireless Communication Dataset: IEEE 802.11ax, LTE, and 5G-NR
Signals~\citep{subray2023wireless} & 192 & 48 & 48 \\
Wireless-technology recognition &
Sub-GHz IQ signals dataset / Multi-band sub-GHz technology recognition
dataset~\citep{fontaine2020subghz} & 384 & 96 & 96 \\
Wireless-technology recognition &
TechRec / Technology Recognition Dataset~\citep{fontaine2019technology,liu2017rssi} & 192 & 48 & 48 \\
\bottomrule
\end{tabular}
}
\end{table}

The modulation recordings contain 128--1,024 samples, the jamming recordings
contain 128--16,368 samples, the wireless-technology recordings contain 4,096
samples, and the UAV recordings contain five million samples.

During the evaluation, each model generates an answer freely. The correctness is
determined using the target labels for the corresponding task. We report
the accuracy for each task and the mean accuracy across the four tasks. Compared
models receive the same questions and label space.

\subsection{LEMMO-LongDesc}
\label{app:datasets_long}

LEMMO-LongDesc addresses long IQ observations and complex
EM scenes. It contains five signal families: 4G LTE, 5G NR,
basic-modulation scenes, ISM-band scenes, and UAV scenes. The dataset provides
targeted QA on specific EM facts and questions requesting descriptions of the complete scene in a long IQ recording. Each signal and all its questions remain in the
same split. Table~\ref{tab:app_long_split} reports the distribution of signals
and questions across the five families.

\begin{table}[htbp]
\centering
\caption{LEMMO-LongDesc composition. The test column reports long-signal detailed description QA.}
\label{tab:app_long_split}
{\setlength{\tabcolsep}{3pt}
\begin{tabular}{@{}lccc@{}}
\toprule
Signal family &
\shortstack{Train\\signals / all QA} &
\shortstack{Validation\\signals / all QA} &
\shortstack{Test signals /\\description QA} \\
\midrule
4G LTE &
\shortstack{2,940 /\\35,280} &
\shortstack{235 /\\2,820} &
\shortstack{120 /\\120} \\
5G NR &
\shortstack{3,675 /\\54,144} &
\shortstack{150 /\\2,225} &
\shortstack{150 /\\150} \\
Basic-modulation scenes &
\shortstack{3,675 /\\25,725} &
\shortstack{150 /\\1,050} &
\shortstack{150 /\\300} \\
ISM-band scenes &
\shortstack{3,675 /\\33,075} &
\shortstack{150 /\\1,350} &
\shortstack{150 /\\150} \\
UAV scenes &
\shortstack{2,138 /\\17,104} &
\shortstack{171 /\\1,368} &
\shortstack{82 /\\82} \\
\midrule
Total &
\shortstack{16,103 /\\165,328} &
\shortstack{856 /\\8,813} &
\shortstack{652 /\\802} \\
\bottomrule
\end{tabular}
}
\end{table}

The training set contains 19,778 detailed description QA pairs and 145,550 targeted QA
pairs. The validation set contains 1,006 description QA pairs and 7,807
targeted QA pairs. The test set contains 802 description QA pairs from 652
distinct physical signals.

The 4G LTE corpus supports recognition of radio-frequency activity and LTE
signals, together with descriptions of observed communication activity. It
covers LTE activity and ambient-noise scenes.

The 5G NR corpus includes questions about NR-signal presence, frequency,
bandwidth, uplink and downlink properties, duplexing, subcarrier spacing,
radio frames, visible terminals or SRS, and resource occupancy.

A basic-modulation observation window may contain multiple component signals.
Its questions address the number of active signals, time-ordered signal
identifiers \(S_1,\ldots,S_n\), modulation types, SNR levels, time--frequency
locations, and overlap relationships. Some reference answers provide both
natural-language descriptions and structured JSON.

The ISM-band corpus covers the number and types of Wi-Fi, Bluetooth, and
related signal sources, as well as protocols, bandwidths, channels, SNRs,
time--frequency overlap, and user counts. The UAV corpus uses DroneRFa
recordings and covers UAV-related signal presence, device models, and scene
descriptions.

The first four signal families are generated with MATLAB; the UAV corpus uses
DroneRFa recordings. IQ recordings in LEMMO-LongDesc contain approximately
0.4--8 million complex samples.

Model inputs comprise the IQ recording, acquisition-setting text, and the user
question. The acquisition-setting text specifies the receiver center
frequency, sampling rate, observation bandwidth, and signal duration. Signal
duration is computed from IQ length and sampling rate.

Evaluation of long-signal descriptions uses free-form generation and assesses the accuracy of specific EM facts in each response. Compared models receive
identical acquisition-setting text, system instructions, and questions. LEMMO
processes each complete IQ recording. Merlin-LoRA supports input sequences of
up to 4,096 IQ samples and therefore processes the first 4,096 samples of each
long recording.

\subsection{EM-134K}
\label{app:datasets_public}
\label{app:datasets_em134k}

EM-134K is a public EM signal--text instruction dataset. Its
released files contain 134,107 QA pairs. Its training and validation splits
support model training and validation-stage model selection.
Table~\ref{tab:app_em134k_split} gives the split statistics.

\begin{table}[htbp]
\centering
\caption{EM-134K split statistics.}
\label{tab:app_em134k_split}
\begin{tabular}{lrr}
\toprule
Split & Distinct IQ signals & QA pairs \\
\midrule
Train & 95,679 & 120,829 \\
Validation & 10,587 & 13,278 \\
\midrule
Total & 106,266 & 134,107 \\
\bottomrule
\end{tabular}
\end{table}

Each record includes fields such as \texttt{id}, \texttt{human\_prompt},
\texttt{gpt\_response}, \texttt{samples}, \texttt{answer},
\texttt{sample\_rate}, and \texttt{snr}. The \texttt{samples} field stores
the IQ recording, \texttt{human\_prompt} contains the question, and
\texttt{gpt\_response} contains the textual response. The \texttt{answer}
field provides the answer key for multiple-choice questions. IQ recordings
contain 512 or 1,024 samples. The dataset covers option-based recognition,
parameter estimation, interference analysis, and strategy generation.

Table~\ref{tab:app_em134k_categories} reports the number of QA pairs in each
content category.

\begin{table}[htbp]
\centering
\caption{EM-134K QA categories.}
\label{tab:app_em134k_categories}
{\setlength{\tabcolsep}{4pt}
\begin{tabular}{@{}p{0.45\linewidth}rp{0.34\linewidth}@{}}
\toprule
Category & QA pairs & Representative objective \\
\midrule
Modulation recognition (\texttt{MOD}) &
9,996 & Identify the modulation type \\
Bandwidth (\texttt{PE\_bandwidth}) &
1,250 & Estimate occupied bandwidth \\
Duty cycle (\texttt{PE\_duty\_cycle}) &
1,250 & Estimate duty cycle \\
Number of pulses (\texttt{PE\_num\_pulses}) &
6,000 & Count radar pulses \\
Pulse repetition frequency (\texttt{PE\_prf}) &
8,500 & Estimate PRF \\
Pulse width (\texttt{PE\_pulse\_width}) &
12,000 & Estimate pulse width \\
Signal-to-noise ratio (\texttt{PE\_snr}) &
7,250 & Estimate SNR \\
First-pulse delay (\texttt{PE\_time\_delay}) &
6,000 & Estimate time delay \\
Protocol recognition (\texttt{PI\_protocol}) &
9,527 & Identify a protocol or standard \\
Fragment detection (\texttt{SD\_fragment\_detection}) &
9,023 & Locate interference-segment boundaries \\
Joint modulation and interference analysis &
6,250 & Identify two signal attributes \\
Interference recognition &
24,437 & Identify interference presence and type \\
Other &
986 & Assess other signal attributes \\
Interference and countermeasure reasoning &
31,638 & Identify interference and generate countermeasures \\
\bottomrule
\end{tabular}
}
\end{table}

EM-134K includes public signals, professional simulations, and real-world
recordings.

\subsection{EM-Bench}
\label{app:datasets_embench}

EM-Bench is a public test benchmark with 14 subtasks and 300 questions per
subtask, yielding 4,200 test questions. Physical signals may be reused across
subtasks; the benchmark contains 3,370 distinct IQ recordings. We use
EM-Bench for testing. Table~\ref{tab:app_embench_tasks} lists the ten
perception subtasks and four reasoning subtasks.

\begin{table}[htbp]
\centering
\caption{EM-Bench subtasks.}
\label{tab:app_embench_tasks}
{\setlength{\tabcolsep}{3pt}
\begin{tabular}{@{}lp{0.40\linewidth}rp{0.30\linewidth}@{}}
\toprule
Capability & Subtask & Questions & Task requirement \\
\midrule
Perception & \texttt{MOD} & 300 & Modulation recognition \\
Perception & \texttt{PE\_Radar\_Bandwidth} & 300 &
Radar bandwidth estimation \\
Perception & \texttt{PE\_Radar\_Duty\_Cycle} & 300 &
Duty-cycle estimation \\
Perception & \texttt{PE\_Radar\_Number\_of\_Pulses} & 300 &
Pulse-count estimation \\
Perception & \texttt{PE\_Radar\_PRF} & 300 &
PRF estimation \\
Perception & \texttt{PE\_Radar\_Pulse\_Width} & 300 &
Pulse-width estimation \\
Perception & \texttt{PI} & 300 &
Protocol or signal-standard recognition \\
Perception & \texttt{CJR} & 300 &
Communication-jamming recognition \\
Perception & \texttt{RJR} & 300 &
Radar-jamming recognition \\
Perception & \texttt{SD} & 300 &
Interference-segment detection \\
Reasoning & \texttt{Anti\_CJ} & 300 &
Generate communication anti-jamming strategies \\
Reasoning & \texttt{Anti\_RJ} & 300 &
Generate radar anti-jamming strategies \\
Reasoning & \texttt{CJS} & 300 &
Analyze and generate communication-jamming strategies \\
Reasoning & \texttt{RJS} & 300 &
Analyze and generate radar-jamming strategies \\
\bottomrule
\end{tabular}
}
\end{table}

The ten perception subtasks use five-way multiple-choice questions, typically
with four specific candidate answers and one ``unable to answer'' option. We
evaluate these subtasks using option accuracy. The four reasoning subtasks
require text-based strategy generation and use ROUGE-L F1. All IQ recordings
contain 512 or 1,024 samples.

\section{Additional Results of LEMMO-LongDesc}

Additional results of LEMMO-LongDesc are shown below.

\begin{figure*}[htbp]
\centering
\includegraphics[width=\textwidth]{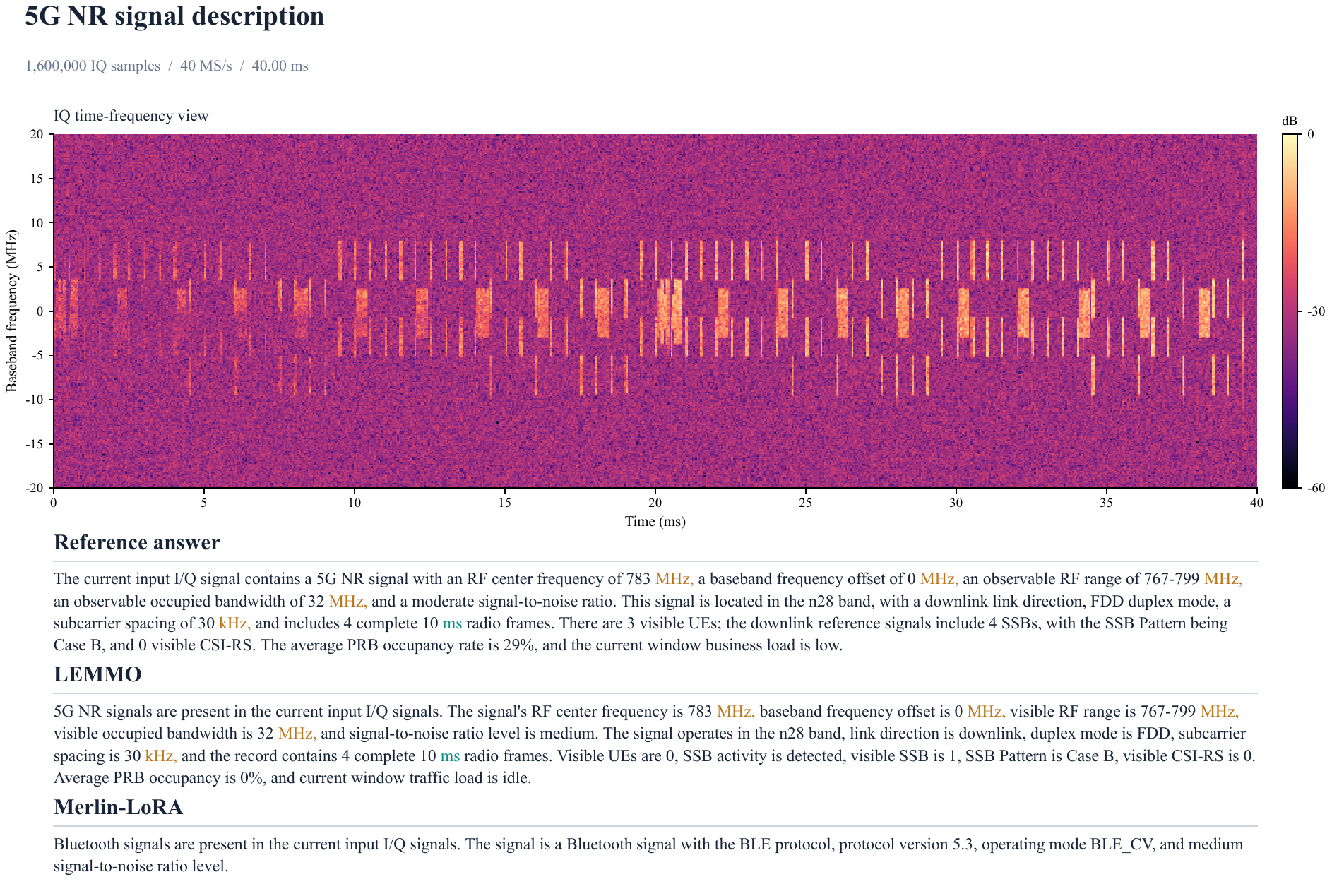}
\caption{Case 1.}
\end{figure*}

\begin{figure*}[htbp]
\centering
\includegraphics[width=\textwidth]{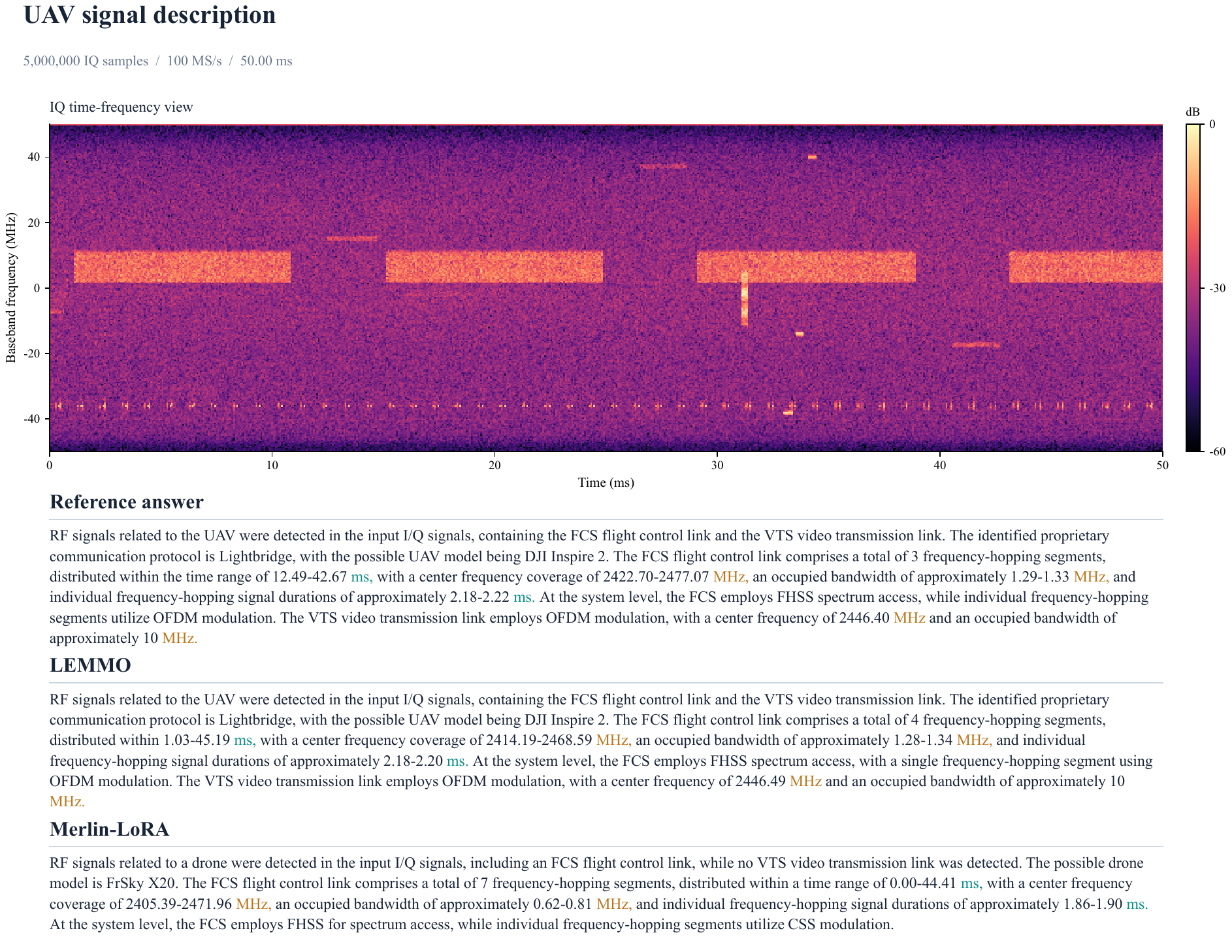}
\caption{Case 2.}
\end{figure*}

\begin{figure*}[htbp]
\centering
\includegraphics[width=\textwidth]{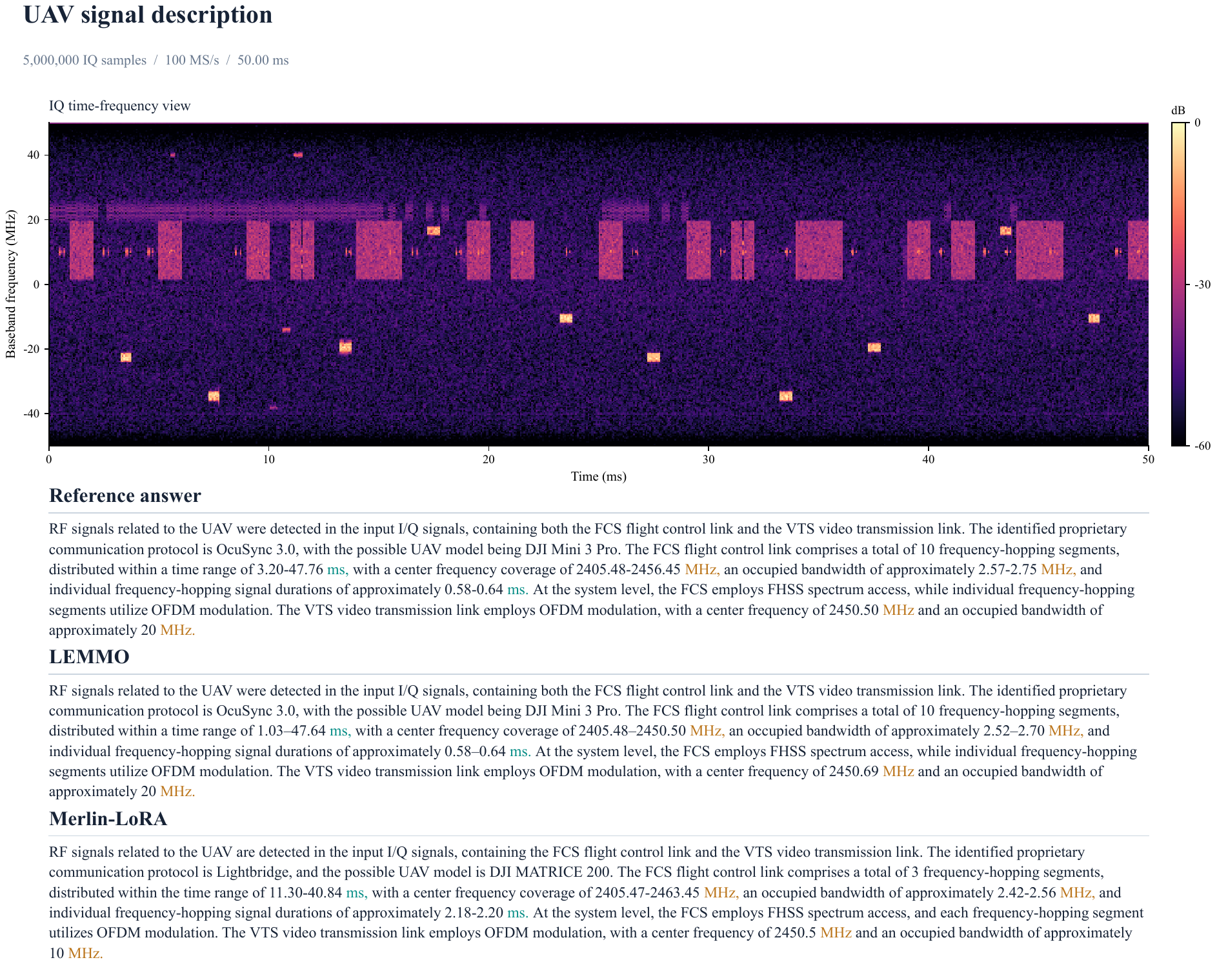}
\caption{Case 3.}
\end{figure*}

\begin{figure*}[htbp]
\centering
\includegraphics[width=\textwidth]{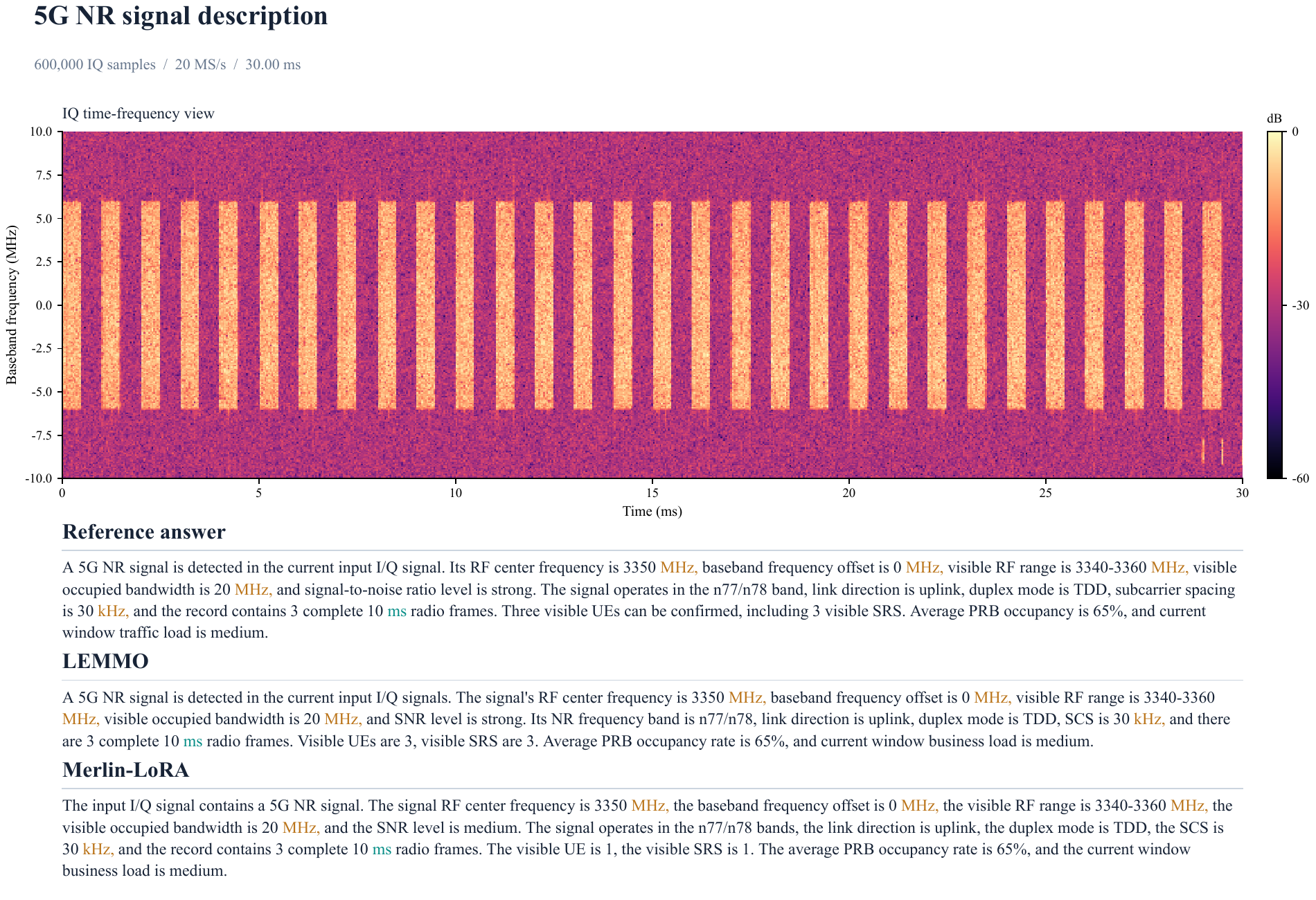}
\caption{Case 4.}
\end{figure*}

\begin{figure*}[htbp]
\centering
\includegraphics[width=\textwidth]{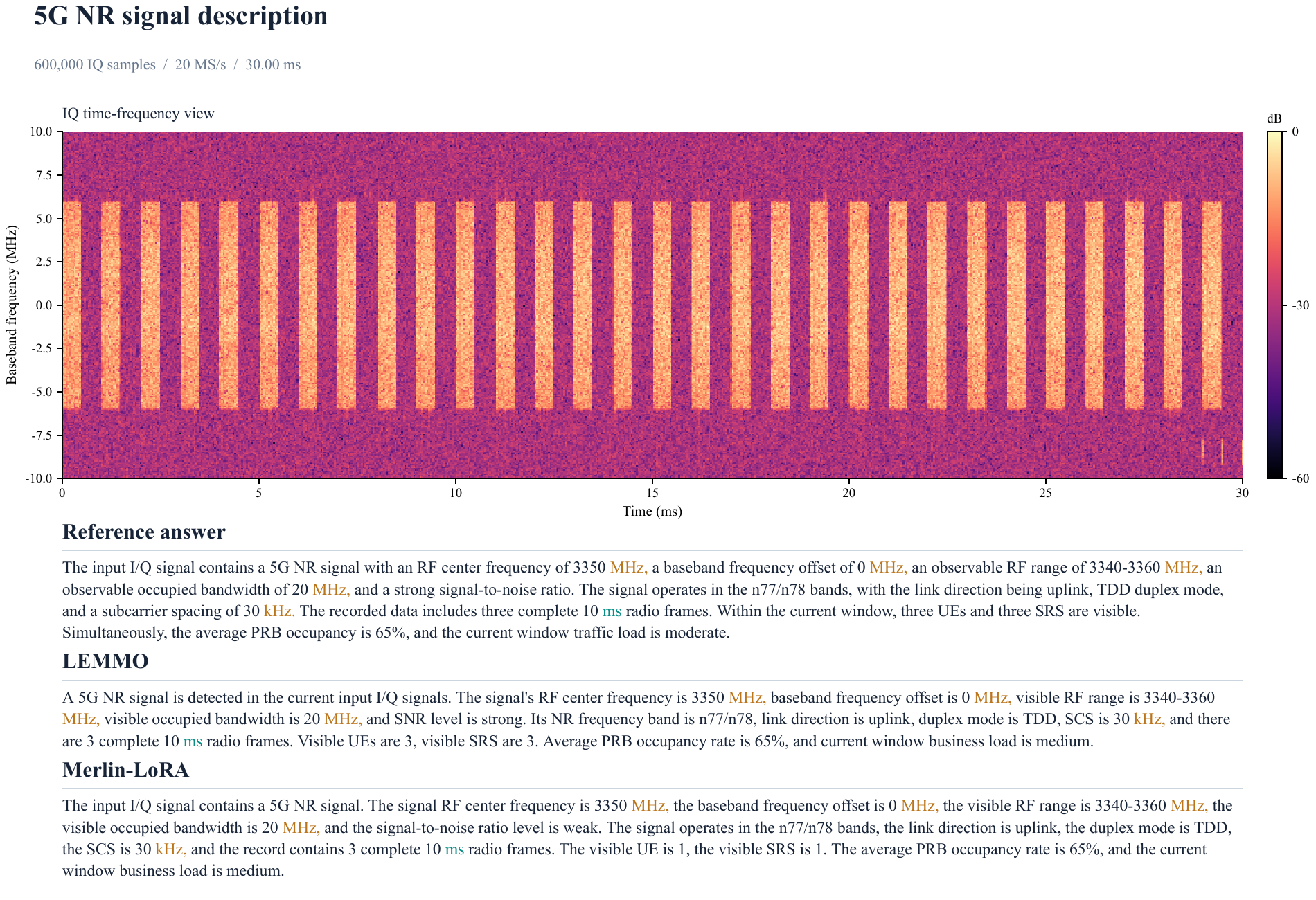}
\caption{Case 5.}
\end{figure*}

\begin{figure*}[htbp]
\centering
\includegraphics[width=\textwidth]{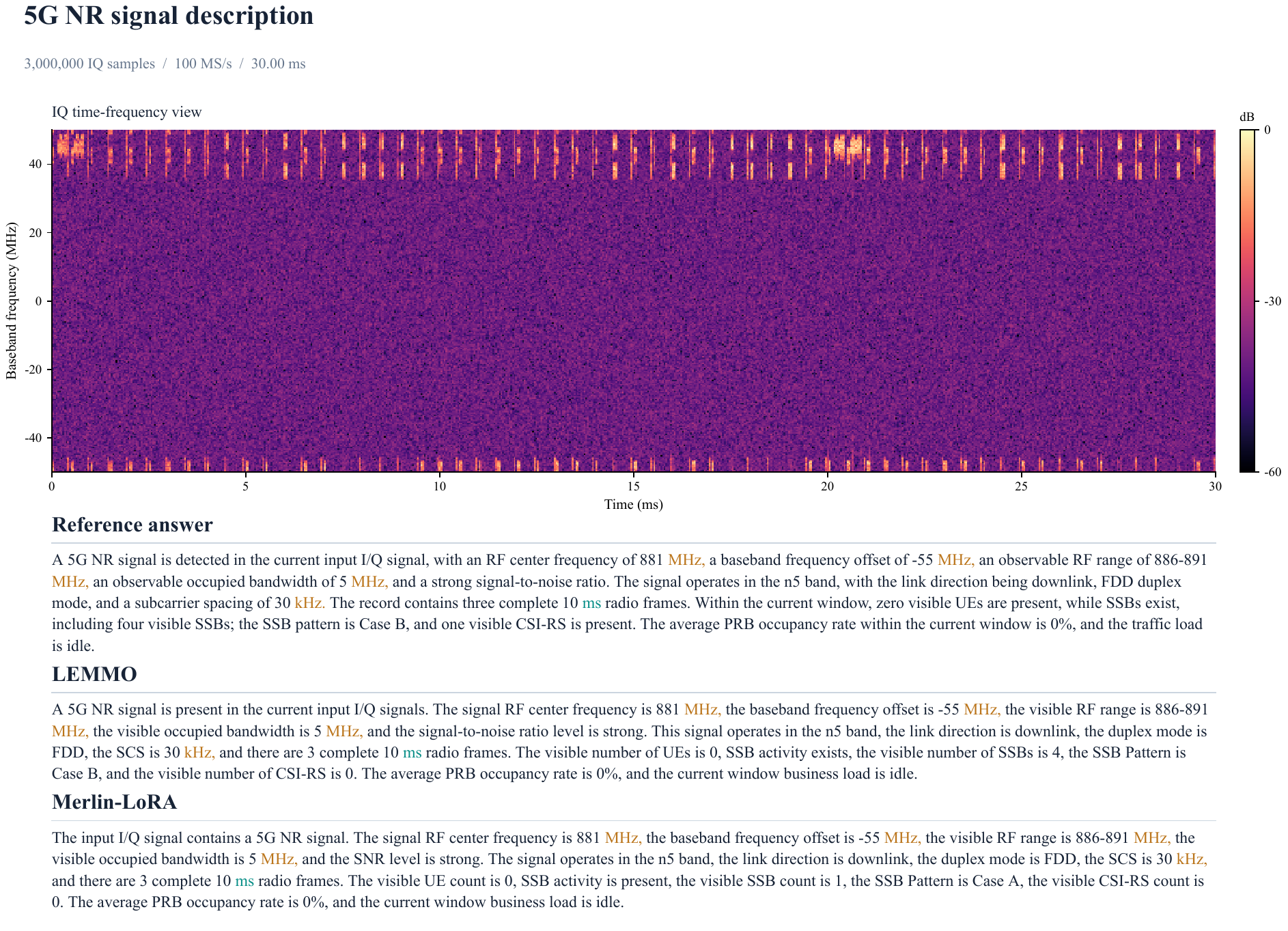}
\caption{Case 6.}
\end{figure*}

\end{document}